# Iterative Feature Exclusion Ranking for Deep Tabular Learning

Fathi Said Emhemed Shaninah[1], AbdulRahman M. A. Baraka[2] and Mohd Halim Mohd Noor[1]

1. School of Computer Sciences, Universiti Sains Malaysia, Pulau Pinang, Malaysia
2. Al-Quds Open University, Palestine

Corresponding email: halimnoor@usm.my

**Abstract**

Tabular data is a common format for storing information in rows and columns, representing data entries and their associated features. Existing deep learning models designed for tabular data often rely on uni-dimensional feature selection mechanisms, which fail to adequately capture contextual dependencies of feature importance. This can result in missed crucial interactions, bias towards high-impact features, and limited generalization in attention generalization. To overcome these challenges, this study introduces a novel Iterative Feature Exclusion module that enhances the estimation of feature importance. The proposed module iteratively excludes a single feature from the input data and computes attention scores that represent the impact of each feature on the prediction. By aggregating the attention scores across iterations, the module generates a refined representation of feature importance that captures both the global and local feature interactions. The effectiveness of the proposed module was evaluated on four public datasets, and the results consistently demonstrated its superior performance over state-of-the-art methods and baseline models in both feature ranking and classification tasks.

The code is publicly available at https://github.com/abaraka2020/Iterative-Feature-Exclusion-Ranking-Module and https://github.com/mohalim/IFENet

## 1 Introduction

Deep learning models for tabular data are widely applied across various domains, such as security [1], academic [2], healthcare [3], e-commerce [4], and fraud detection [5]. The success of this approach largely depends on the model's ability to accurately identify the most influential features among all predefined features in the dataset. The task of identifying relevant features is called feature selection. One of the common approaches in feature selection is feature importance ranking (FIR), a technique that ranks features based on their contribution to the model's predictive performance. Specifically, FIR refers to the task of determining the impact of each input feature on the performance of a supervised learning model [6]. This is achieved by computing an importance score for each feature, indicating its contribution to the model's decision-making process. A high score corresponds to a greater impact on the model's performance and vice versa. Based on the importance score of the features, an optimal subset of features can be obtained by selecting those with high scores and dropping those with low scores.

The attention mechanism is a state-of-the-art technique used in deep learning models that assigns weights to features, reflecting their relevance to the model's predictions. The weights are computed based on the dependencies between features and the current context or input, allowing the model to focus on the most informative part of the data. Attention mechanisms typically assign a weight to each feature based on its contribution when all features are simultaneously present, indicating its overall importance. Although effective for highlighting globally dominant features, this approach tends to overemphasize strong predictors and underrepresent context-dependent or interactions between features. In addition, this approach might not capture the nuanced way in which features interact and influence outcomes in different contexts. Thus, it may overlook the differences in feature importance that are crucial for complex tasks due to its weakness in identifying feature interactions with each other. On the other hand, traditional attention can sometimes suppress the contributions of certain features, potentially leading to inaccurate classification results. For example, consider credit risk prediction, attention mechanisms may assign a larger score to the income feature and a lower score to the debt-to-income ratio feature due to the strong correlation between income and credit risk. However, the debt-to-income ratio may receive a higher score when it is evaluated in isolation (without the presence of the income feature), as its individual contribution to predicting credit risk becomes more significant due to the absence of overlapping and redundant information provided by the income feature.

To address this limitation, we introduce IFENet, a deep tabular model integrated with an innovative Iterative Feature Exclusion (IFE) module designed to enhance the understanding of feature importance in tabular data. The proposed module systematically evaluates the relevance of individual features by iteratively excluding one feature from the input data and assessing its effect on the resulting attention scores. In each iteration, the proposed module captures the importance of the remaining features in the absence of the excluded feature, referred to as local feature importance. The local feature importance reflects the relationships and interactions among the subset of features in each iteration. After completing all iterations, the module aggregates the local importance scores to produce a comprehensive view of feature interactions, referred to as global feature importance. This global feature importance offers a more refined understanding of feature relevance, incorporating both local interactions and global dependencies across the feature set.

The proposed method offers the following contributions:

- Development of an innovative iterative feature exclusion module to enhance the feature ranking method for tabular data.
- Enhanced estimation of feature importance, leading to improved classification model performance.
- Improved interpretability: By iteratively excluding features and observing changes in attention scores, individual feature contributions can be better understood.

The next section reviews related work on feature importance ranking and deep learning models for tabular data, identifying current limitations. Then, we present the architecture and operational principles of the proposed module in detail. Finally, a comprehensive experimental evaluation of the proposed method is provided across various benchmark datasets, along with discussions and insights into its performance.

## 2 Related Work

Broadly, tabular data is a collection of feature values, and modeling such data often requires learning both the importance of these features and their relationships. FIR determines not only which features are relevant or irrelevant but also the degree of feature importance on the final output. FIR methods are generally divided into three categories: filter, wrapper, and embedded methods. Filter methods rank the features using various selected metrics such as statistical measures, mutual information, variance, and correlation scores. They offer computational efficiency and operate independently of particular machine learning algorithms. Analysis of Variance (ANOVA) is a filter statistical approach that uses mean differences among different groups of scores to select the features [7]. The Spearman's rank correlation is another filter method that assesses the relationship between two variables based on their rankings rather than their actual values, applicable when data is not normally distributed [8]. However, filter methods do not capture interactions between features [9].

Wrapper methods involve training and evaluating a machine learning model multiple times to evaluate the quality of feature subsets based on their actual model performance [10]. These methods are not model-independent and can therefore be based on any model. Forward selection, backward elimination, and recursive feature elimination are widely used techniques. Recursive Feature Elimination (RFE) excludes the least influential features iteratively and reassesses model performance at each iteration. Starting with the full feature set, it progressively discards those contributing the least—based on measures such as feature coefficients or importance scores—until identifying an optimal subset of features [11]. In addition, Petković, et al. [12] propose a method for semi-supervised learning of feature rankings that consists of two feature ranking approaches. In the first approach, a ranking is computed from an ensemble of predictive clustering trees for structured output prediction. The ensemble-based feature ranking is based on three ensemble learning methods combined with scoring functions. The second approach is the Relief method which determines the relevance of a feature by comparing the values of the feature between a given instance and its neighbouring instances from the same and different classes. Although wrapper methods can improve model accuracy, they tend to be computationally demanding and may not scale efficiently with large datasets.

Embedded methods integrate a feature selection method directly into the process of training a machine learning model. Examples include tree-based, and neural-network-based methods. In tree-based methods, feature selection occurs during model training. These methods assess feature importance based on their contributions to decision splits, whereby features that reduce impurity or appear more frequently are considered more important, while less relevant features are implicitly neglected. Common examples include decision trees [13], random forests [14], and gradient boosting [15]. These methods are computationally efficient, as feature selection is seamlessly integrated into the training process rather than performed as a separate, iterative step. However, tree-based methods are non-differentiable and incompatible with gradient-based optimization; thus, they cannot be seamlessly integrated into end-to-end trainable architectures [16]. In addition, traditional decision trees and ensembles like Random Forests often rely on greedy, axis-aligned splits. This can bias feature importance and limit their ability to model intricate, distributed interactions between features [17]

Recently, numerous neural network-based methods have been proposed to enhance feature selection by leveraging the model's ability to learn complex patterns and representations directly from data. The methods integrate feature selection into the learning process, aiming to enhance interpretability and generalization performance. In general, these methods can be divided into non-attention-based methods and attention-based methods. Joseph and Raj [18] propose a Gated Adaptive Network for Deep Automated Learning of Features for Tabular Data (GANDALF) and a Gated Feature Learning Unit (GFLU) that integrates feature selection and gating mechanisms inspired by Gated Recurrent Units (GRU). The model processes the input data by a series of hierarchical formulations involving multiple stages using GFLU units, using weights unique to each stage. In addition, a learnable mask is constructed by applying a sparse transformation on a learnable parameter vector. At each stage, a learnable mask is used for the soft selection of important features of feature learning in the GFLU. The feature selection through masks is exclusively employed for the gates. The model aims to ensure efficient and interpretable learning by selectively focusing on important features at each stage, thereby improving performance and computational efficiency. Primožič et al. [19] propose an unsupervised feature ranking algorithm based on a graph network. The model attempts to reconstruct a representative network of features by ranking nodes in this network via the efficient PageRank approach.

Attention-based methods leverage attention mechanisms in their feature selection method, such as self-attention, network attention, or multi-head attention. Škrlj et al. [20] investigate the estimation of feature importance to explain models trained on propositional tabular data called self-attention network (SAN). Feature importance can be directly estimated from the network's internal attention mechanisms by averaging the attention vectors across all training instances. Furthermore, the averaging can be performed over only correctly classified instances, providing more reliable explanations and reducing noise. Finally, the global feature importance vector can be derived directly from the learned attention weights using the SoftMax function on the diagonal elements of the attention weight matrix. This study provides empirical evidence that SANs can produce attention vectors that provide feature rankings comparable to those of an external classifier. Zhang et al. [21] introduce a feature weighting approach for tabular data. The reinforcement learning strategy is used to fine-tune the feature weights based on feedback from downstream tasks. The proposed approach consists of three components: feature alignment, feature weighting, and fine-tuning. Feature alignment aligns different types of original features so that they exist a shared representation space. In the feature weighting stage, the feature matrix is first encoded to get its embedding using transformer encoders, and then the embedding is decoded into feature weights. The transformer encoders use multi-head self-attention to capture feature relationships, followed by a cross-attention mechanism to generate a contextual representation that captures the dependencies within the feature matrix. In the fine-tuning stage, a reinforcement learning strategy is employed to adjust the feature weights based on feedback from downstream tasks.

Liu et al. [22] propose a novel method known as discovering instance-wise influential features for modeling tabular data (DIWIFT). Rather than selecting a set of features based on all instances, DIWIFT uses a self-attention network to identify and choose influential features on an instance-by-instance basis. DIWIFT follows a four-phase approach. First, it trains a base deep learning model on unaltered tabular data to obtain initial parameters, which are used in subsequent computations. Second, a self-attention-based feature selector estimates per-instance probabilities to determine which features should be retained. Third, these filtered instances are fed into an influence function module that computes the impact of selected features. Finally, these influence scores are used to shape a custom loss that refines the feature selector through backpropagation, thereby improving instance-wise feature selection. Prabha et al. [23] propose an attention-based model that explicitly generates feature-level attributions as part of the prediction process. The model learns and assigns interpretable importance weights to each feature per instance. Specifically, the model embeds each input feature using either a shared or feature-specific linear layer, producing vectors that serve as keys and values for an attention mechanism. The global query vector is defined from learned parameters or derived from contextual information. Each feature's relevance is computed via a dot product between the query and its corresponding key, and the resulting scores are normalized using softmax function to produce. These attention weights are then used to combine the feature values into a single aggregated representation. This representation is passed to a classifier, and the attention weights themselves serve as interpretable indicators of each feature's contribution, making the model behave like an adaptive linear model for each instance. Furthermore, Xue et al. [24] developed an approach that uses attention units within a neural network to rank features based on their impact on accuracy. Their findings highlighted the importance of attention in identifying relevant elements, especially for high-dimensional data. However, they noted the shortcomings of the method, such as its reliance on data quality and the risk of overlooking variables such as interpretability. Furthermore, Yasuda et al. [25] explore sequential attention, providing a strategy that considers the residual value of features or their contribution in addition to those already chosen. This approach proved effective in capturing the dynamic relationships between characteristics. However, its sequential nature may not be appropriate for all datasets.

Embedded methods offer a trade-off between computational efficiency and the model's performance, which contributes to their widespread adoption across various applications. Most attention-based feature ranking models assign a single weight to each feature, representing its overall importance across the entire feature set [20], [22]. However, this approach may overlook the subtle interactions between features and their varying influence across different situations. It may also fail to fully capture the complex dependencies or provide a clear and intuitive understanding of feature importance, albeit offering some interpretability by emphasizing essential key aspects. In addition, Guo et al. [26] proposed an iterative algorithm for multi-label feature selection that enhances the reliability of instance similarities by integrating local and global similarity structures within a reduced feature space. The proposed method employs a two-stage iterative learning method that progressively refines instance similarities and mitigates the influence of irrelevant features. In the first stage (W-stage), feature weights are learned by integrating pseudo-labels into a regression model, constrained by both local manifold structures and global instance similarities using dual graph regularization. In the second stage (S-stage), instance similarities are dynamically updated within the reduced feature space based on the selected features from the previous iteration, allowing the model to iteratively refine both the feature selection and similarity structure for improved accuracy and robustness in multi-label tasks.

Furthermore, some studies directly quantify instance-dependent feature importance and linear weighted aggregation, which still fail to capture the complex relationship between features [23]. On the other hand, transformer-based models typically require large amounts of training data to learn consistent attention patterns. In low-data or highly imbalanced datasets, these models may struggle to assign reliable weights and are prone to overfitting, potentially degrading performance [21].

In summary, while embedded methods offer a powerful trade-off between performance and integrated feature selection, they typically assign a single, static importance weight to each feature, computed from the entire feature set. This approach risks overlooking nuanced, context-dependent interactions, as the presence of dominant features can suppress the importance of correlated or contextually relevant ones. Consequently, they may fail to capture the change in a feature's importance based on the presence or absence of others. Similarly, RFE and other wrapper methods are inherently computationally expensive due to their need to train and evaluate a model multiple times from scratch. This iterative retraining process makes them impractical for integration within end-to-end deep learning architectures and limits their scalability. In addition, wrapper feature selection methods remove features iteratively based on their importance score; thus, the risk of the effect of dominant features still exists.

This study presents IFENet, a deep learning model specifically designed for tabular data, which incorporates a mechanism for feature ranking to enhance interpretability and identify the most relevant predictors. IFENet integrates a novel Iterative Feature Exclusion (IFE) module that systematically excludes one feature at a time and evaluates the importance of the remaining features in contributing to the prediction. The importance scores obtained from each iteration were then aggregated to produce an overall feature importance score. The proposed model can better identify and understand nuanced interactions and dependencies between features.

## 3 Proposed Methodology

This section presents IFENet, a deep learning model designed to handle tabular data that incorporates the proposed IFE module, which captures complex feature interactions to enable feature ranking and identify the most relevant features of the data. The architecture IFENet incorporated with the proposed IFE is illustrated in Figure 1. As shown in Figure 1, the proposed module consists of masking operations, feature attention units, and stacking and aggregation. In general, the feature importance score is calculated iteratively using the feature attention unit. In each iteration, the module excludes a single feature and measures the attention scores of the remaining features. The proposed module focuses on the input features and generates attention scores that highlight the relative importance of the features within the context of the current iteration. This process enables the model to capture local patterns, i.e., context-dependent interactions and importance relationships that may vary depending on the subset of features present in each iteration.

The attention scores for all iterations are then stacked and aggregated to obtain the feature importance scores. The feature importance scores represent the extent to which each feature contributes (or the utility) to the model's prediction. It determines the degree of usefulness of the input features for a classification (or regression) model output. By aggregating the attention scores from all iterations, the model constructs a global view of feature importance, identifying features that consistently contribute to predictions across diverse contexts. This aggregation allows IFENet to recognize both universally relevant features and those whose importance is conditional, resulting in a refined and robust representation of the feature importance. A higher score has a larger effect on the prediction, and vice versa. This provides a more nuanced understanding of feature importance by evaluating the impact of excluding each feature individually. This allows for a detailed analysis of how each feature contributes to the model performance and more accurate interpretations of feature importance.

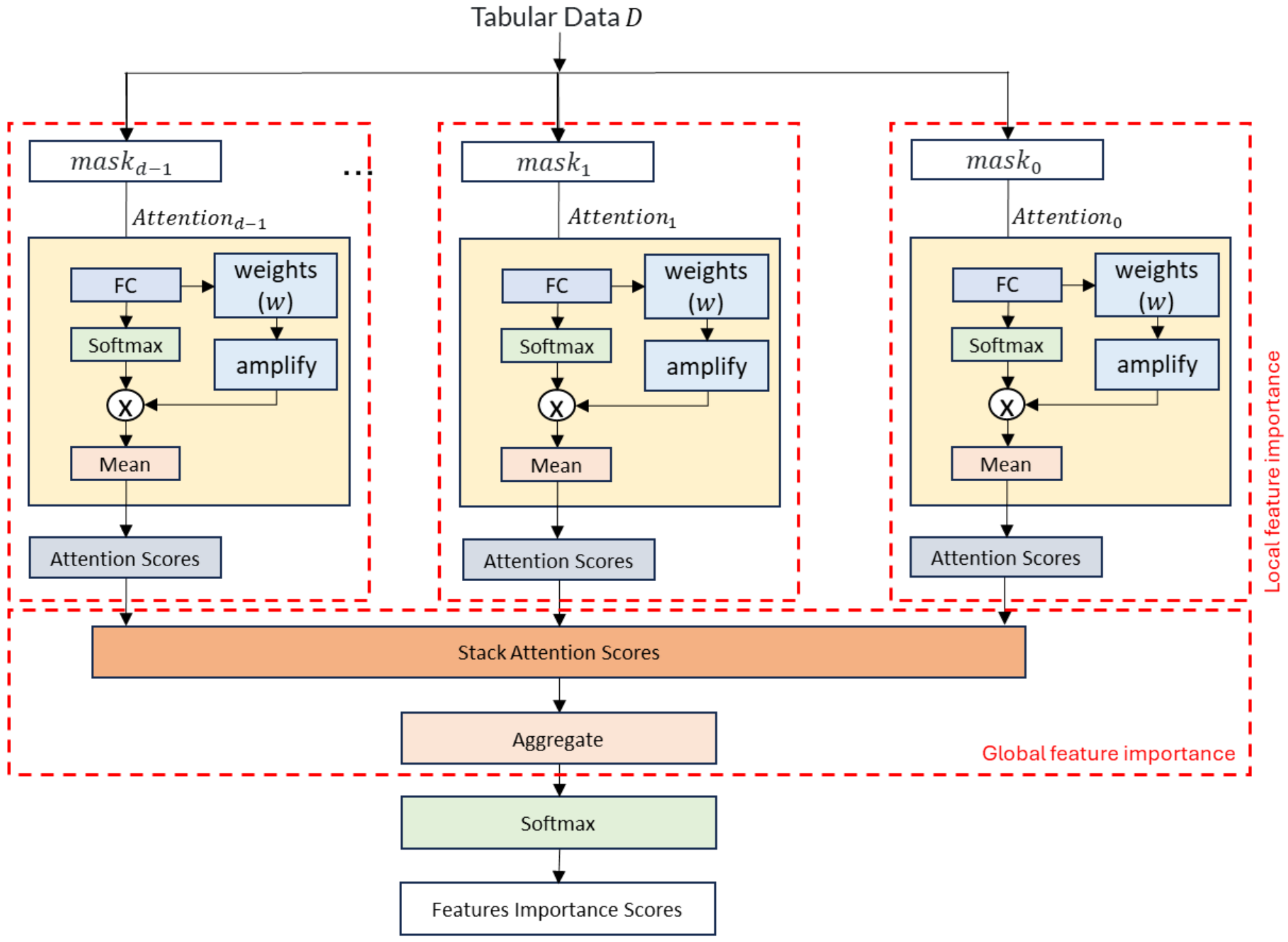

(a) The architecture of the proposed IFE module

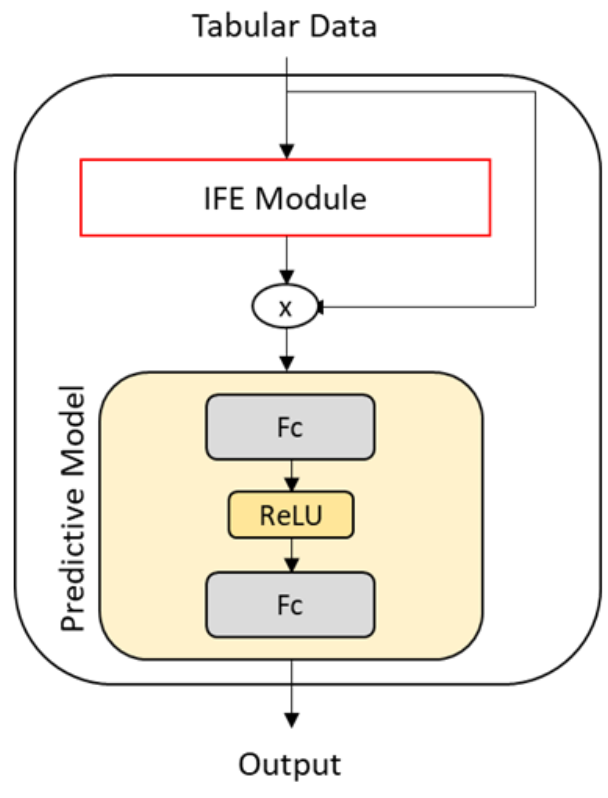

(b) The architecture of IFENet

Figure 1. Architecture of the proposed IFE module and IFENet.

### 3.1 Iterative Feature Exclusion Module

Initially, the IFENet accepts raw tabular data $\boldsymbol{X}^i$ from dataset $D = \{(\boldsymbol{X}^1, y^1), \ldots, (\boldsymbol{X}^n, y^n)\}$, $\boldsymbol{X}^i \in \mathbb{R}^d$, where $n$ refers to the number of instances and $d$ refers to the number of features. For classification problems, each instance $X^i$ is associated with a class label $y^i \in \{0,1, \ldots, C-1\}$ where $C$ denotes the number of target classes. For regression problems, $y^i \in \mathbb{R}$ is a real-valued target.

Next, the IFE iteratively excludes a single feature using a binary mask. The attention scores of the features are then calculated using the attention unit. Therefore, the module builds $d$ number of attention units, which is equal to the number of masks.

This iterative exclusion mechanism is key to understanding the nuanced impact of these features. By systematically excluding one feature at a time and re-evaluating the attention scores of the remaining features, the module can discern not only the individual importance of each feature (local interaction) but also how the importance of features shifts in the context of other features being absent or present (global interactions). For example, if two features are highly correlated, removing one might significantly increase the perceived importance of the other, revealing their contextual dependency on each other. This process allows the model better to predict the contribution of each feature in various configurations, leading to a more robust and comprehensive understanding of feature importance than a traditional attention mechanism, which might overlook such interdependencies.

Specifically, in each iteration, the masking operation creates a binary feature mask $\boldsymbol{m}$, a vector to indicate which feature is being excluded, $\boldsymbol{m} = (m_1, \dots, m_d)$, where each element $m_j \in \{0,1\}$. The vector is used to exclude a single feature sequentially, i.e., the $j^{th}$ feature is excluded by assigning a zero value (masking) for the $j^{th}$ feature. In this case, the number of iterations is equal to the number of features $d$. Let $\widetilde{\boldsymbol{X}}$ the masked input data, the masking is performed by multiplying each feature element-wise with the corresponding element in the mask vector.

$$\widetilde{\boldsymbol{X}} = \boldsymbol{X}^i \times m_j \tag{1}$$

The attention unit then accepts the masked data to calculate the attention scores of the features. The attention scores reflect how each feature influences the network outcome. The attention unit consists of a fully-connected (FC) layer followed by the SoftMax function. The number of units in the FC layer is equal to the number of classes $C$. The SoftMax function is used to convert the output of the FC into a probability distribution, where each value is between 0 and 1, and the total sums to 1.

The attention scores $\boldsymbol{z}_j$ for $j^{th}$ iteration are obtained by computing the FC layer with SoftMax on the masked data $\widetilde{\boldsymbol{X}}$ as follows:

$$\boldsymbol{z}_j = g\left(\boldsymbol{w}_j \cdot \widetilde{\boldsymbol{X}}\right) \tag{2}$$

Where $\boldsymbol{w}_j$ is the weight matrix of the fully-connected layer for $j$-th iteration and $g$ is the SoftMax function, $\boldsymbol{w}_j \in \mathbb{R}^{d \times C}, \boldsymbol{z}_j \in \mathbb{R}^C$.

The output of the unit is a vector consisting of the attention scores, $\boldsymbol{a}_j$ for each feature. The dot product between the result of the SoftMax function and the "amplified" weights is calculated. The weights are amplified using an amplification coefficient $r$, which is a hyperparameter that controls the sensitivity of the weight amplification. Multiplying by the hyperparameter increases the positive weights, thereby allowing significant features to stand out. In contrast, negative weights become smaller, suppressing the less important or detrimental features. The weights are amplified by multiplying them by the $r$ and applying an exponential function as follows:

$$\boldsymbol{a}_j = \mathrm{e}^{\boldsymbol{w}_j \circ r} \odot \boldsymbol{z}_j \tag{3}$$

where $\odot$ is an element-wise multiplication. The amplification coefficient widens the range between the positive and negative weights and helps manage the gradient magnitudes during backpropagation, leading to a more stable training process. Given that the optimal value of $r$ can vary by dataset; empirical tuning is recommended. Based on our experiments, a promising initial search range for $r$ is typically between 1.0 and 6.0, as peak performance values were observed within this interval across the tested datasets. It is important to note that while the performance stabilizes beyond $r$=6.0, lower values within the optimal range often provide a better balance between sensitivity and stability.

However, the differences between the values should maintain their relative distance. A large value of $r$ leads to high differences, whereas a small value diminishes their impact. By appropriately choosing the value of the constant, we can amplify or attenuate the weights as needed. This allows the module to focus more strongly on the relevant weight values within the matrix.

The proposed module computes attention scores for each iteration, which refers to local feature importance, where it considers different feature subsets for the instance. The calculated attention scores from each iteration help avoid the effect and bias of high-impact features on the other features.

The attention scores from each iteration are stacked together to obtain the attention scores matrix $\boldsymbol{A} \in \mathbb{R}^{dxd}$. The concatenated attention scores provide an accurate representation of feature importance, effectively capturing the global feature importance.

$$\boldsymbol{A} = \text{concatenate}(\boldsymbol{a}_1, \boldsymbol{a}_2, \dots \boldsymbol{a}_d) \tag{4}$$

The concatenated attention scores are then averaged to reduce the overly influential impact of extreme values. Finally, the feature importance scores $\boldsymbol{S} = \{s_1, s_2, \dots s_d\}$ are obtained by applying the SoftMax function to the average scores. By combining the attention scores from each iteration, the model generates a more refined representation of feature importance, capturing both global and local patterns in the tabular data. The feature importance scores are computed by

$$\boldsymbol{S} = g(\text{mean}(\boldsymbol{A})) \tag{5}$$

Mean aggregation provides a straightforward and transparent way to observe which features are consistently important across iterations and provides a simple and interpretable representation of global feature importance. $\boldsymbol{S}$ vector presents the feature importance scores of all the features. Each value represents the importance of the corresponding feature. i.e., high value refers to high importance and vice versa. Since aggregation is essential to maintaining the interpretability of feature importance scores over the iterations, the mean method is used to minimize variations in the attention and balance contributions across all iterations. Mean aggregation gives a simple and clear justification of which features are consistently significant during iterations and provides a consistent aggregation method across various datasets.3.2 Predictive Layers

A fully connected neural network is used as the predictive layers, as shown in Figure 1(b). The network can approximate increasingly complex decisions and achieve high prediction accuracies by increasing the number of hidden layers, i.e., the depth of the network[27]. In this study, the predictive architecture consists of two fully connected layers, representing the hidden layer with ReLU and the output layer. The hidden layer size is assigned based on the number of features. The layers and their parameters are shown in Table 1.

The raw tabular data is passed to the proposed module. To ensure stable training and better convergence, we apply batch normalization to the input data. The IFENet accepts the normalized data and then measures the feature importance scores, $S$ for each feature. The feature importance scores represent the importance of each feature. The feature importance score is then used to generate weighted input data for the predictive model as follows:

$$\boldsymbol{Z} = \boldsymbol{S} \odot \boldsymbol{X} \tag{6}$$

The element-wise multiplication gives more weight to the relevant features and less weight to the irrelevant features.

It is worth noting that, unlike the feature selection process, the proposed module assigns an importance score to all features without excluding or filtering any features. The weighted input is then fed into the predictive layers, which consist of a hidden fully connected layer with ReLU activation, followed by an output fully connected layer to produce the final prediction.

Table 1. Architecture of the predictive layers.

| | Layer | Parameters and values |
|---|---|---|
| 1 | Fully connected | units = $d$, activation = ReLU |
| 2 | Fully connected | units = $C$ |

This allows the model to obtain a more comprehensive understanding of feature importance, enhancing the model's ability to capture complex relationships within the data and understand how each feature contributes to the overall attention.

### 3.2 Theoretical Justification and Methodological Advantages

The proposed IFENet is theoretically grounded in the principle that feature importance is context-dependent rather than absolute. In tabular learning, a feature’s contribution is not fixed; it varies according to its interactions and correlations with other features. Traditional attention mechanisms compute feature weights in a single forward pass, producing static importance scores that often overlook conditional dependencies and feature masking effects.

IFENet introduces an iterative exclusion mechanism, where one feature is temporarily excluded at each iteration and the attention distribution is recalculated. This exclusion acts as a context-aware perturbation, compelling the attention unit to redistribute relevance among the remaining features. Through this iterative process, IFENet effectively measures the sensitivity of the model to the absence of each feature impact, yielding a richer estimation of its importance.

Mathematically, let $\boldsymbol{a}_j$ denote the attention distribution at iteration $j$ and $d_j$ represent the feature excluded at $j$ step. The shift in attention can be captured by:

$$\Delta \boldsymbol{a}_j = \boldsymbol{a}_j - \boldsymbol{a}_{j-1} \tag{7}$$

This represents how the exclusion of $d_j$ alters the importance of the feature. Aggregating this attention across all iterations provides a global-context representation that reflects not only individual contributions but also their interaction effects.

The proposed IFENet framework provides several methodological advantages over traditional attention-based and elimination-based approaches. Revealing features' importance: by iteratively excluding dominant features, IFENet uncovers features whose relevance is previously overshadowed in a full-feature context. This enables the model to capture latent dependencies that may be ignored by a single pass. Capturing complex interactions: each exclusion iteration recalibrates the attention distribution, implicitly modeling feature co-dependencies. This mechanism allows IFENet to approximate interaction effects without explicitly enumerating feature combinations. Differentiable and efficient evaluation: unlike RFE, which requires retraining after each removal, IFENet performs exclusion internally within a single end-to-end forward process. This ensures computational efficiency and gradient-based optimization compatibility. And finally, it's providing context-sensitive ranking, where IFENet produces attention-based rankings that dynamically adjust according to the local feature context, offering a more nuanced interpretation of feature relevance under varying conditions.

## 4 Experiments and Results

To comprehensively evaluate the performance of the proposed module, four different experiments are performed. The first experiment aims to evaluate the impact of the proposed module on the performance of prediction tasks, whereas the second experiment aims to evaluate the ranking of feature importance. The proposed module is compared with the baseline in terms of classification in the third experiment, while the last experiment aims to evaluate the proposed module with state-of-the-art models in terms of classification.

### 4.1 Datasets

Four public real-world classification datasets are used in experiments. The datasets include Telco Customer Churn (TELCO)[28], Home Equity Line of Credit (HELOC)[29], Titanic [30], Adult [31] , and Student Performance (UCI-STP) [32] which are adopted in previous studies focusing on modeling tabular data. For a fair comparison, all models have the same train-validation-test splits. The split ratio is approximately 75%, 5%, and 20% for training, validation, and testing, respectively, as shown in Table 2.

### 4.2 Experimental Setup

Since tabular data is heterogeneous data (i.e., contains numerical and categorical data), two main preprocesses are performed for all datasets. Removing missing values and encoding the categorical features by one-hot encoding. No normalization method is used for numerical data.

Table 2. Datasets summary

| Dataset | Features | Instances | | |
|---|---|---|---|---|
| | | Train | Validation | Test |
| TELCO | 20 | 5274 | 88 | 1670 |
| HELOC | 23 | 7844 | 522 | 2093 |
| Titanic | 14 | 781 | 66 | 195 |
| UCI-STP | 32 | 783 | 52 | 209 |
| ADULT | 15 | 22621 | 1508 | 6033 |

The parameters for the proposed module, including learning rate, hidden size, and batch size, are optimized by search spaces for all datasets, as illustrated in Appendix A.1. The maximum epoch is 120.

The hyperparameters of state-of-the-art models are fine-tuned by Bayesian Optimization (BO) using the Optuna library[33]. The number of trials is 50. The early stop is used for validation accuracy with 10 patience. The list of corresponding models' hyperparameters, along with their respective search spaces, is listed in Appendix A. The default value is used for any other parameter not listed.

All experiments are implemented in Python. The state-of-the-art models are implemented using PyTorch tabular library [34], and the proposed method is implemented using PyTorch lib on a workstation equipped with Intel Core i5 10400f CPU, 64G RAM, and GPU GTX 1660 Super. In addition, for each dataset, the same preprocessing was used for all models

The evaluation of the model performance is performed in terms of accuracy, average precision ($P$), average recall ($R$), and average f-score ($F$), which is calculated through equations 7-9:

$$accuracy = \left(\frac{(TP+TN)}{TP+FP+TN+FN}\right) \tag{8}$$

$$P = \frac{TP}{TP+FP} \tag{9}$$

$$R = \frac{TP}{TP+FN} \tag{10}$$

$$F = 2 \times \frac{P_c \times R_c}{P_c + R_c} \tag{11}$$

where $TP$ is a true positive, $TN$ is a true negative, $FP$ is a false positive, $FN$ is a false negative, and $c$ is the class label.

In addition, we evaluate the feature importance ranking using normalized discounted cumulative gain@K (NDCG@K) [24]. NDCG@K values range from 0 to 1, where 1 indicates the ideal ranking, and 0 indicates the worst possible ranking.

## 4.3 Evaluation and Discussion

### 4.3.1 Ranking of Feature Importance and Interpretability

To assess the effectiveness of the proposed module in ranking important features and enhancing interpretability, we conduct a series of experiments focused on feature ranking evaluation.

Since direct ground-truth feature rankings are typically unavailable for tabular datasets, we adopt GradientSHAP [35] to approximate the ideal feature importance distribution. SHAP remains a robust interpretability technique for deep neural networks, making it a suitable reference in this context [36].

GradientSHAP leverages SHAP values to quantify each feature's contribution to the model's prediction and is considered a reliable method in terms of feature importance ranking, and is interpretable in related studies [36], [37].

To evaluate the effectiveness in terms of feature ranking and interpretability of our proposed IFENet module, we compare its feature ranking performance against XGBoost (as a baseline) using the Normalized Discounted Cumulative Gain (NDCG@K) metric at various top-K positions ($K \in$ $\{1, 2, ..., d\}$), which measures how closely the ranked list of features from a model aligns with the reference ranking. First, the Discounted Cumulative Gain (DCG) for the top-K ranked features is calculated as:

$$DCG@K = \sum_{i=1}^{K} \frac{GS_i}{\log_2(i+1)} \quad (12)$$

where $GS_i$ is the graded relevance (i.e., the GradientSHAP value) of the feature at position $i$. The NDCG@K is then computed by normalizing the DCG@K by the Ideal DCG@K (IDCG@K), which is the DCG@K of the perfect ranking:

$$NDCG@K = \frac{DCG@K}{IDCG@K} \quad (13)$$

A higher NDCG@K score (closer to 1.0) indicates better alignment and, therefore, stronger interpretability, and a lower score from 1.0 refers to lower performance.

The results in Figure 2 show that the NDCG@K scores of IFENet are consistently closer to 1.0 than those of XGBoost across all values of K. This indicates that IFENet produces rankings more aligned with SHAP-based importance. This suggests that IFENet yields better interpretability in terms of identifying the most relevant features.

The results show that, overall, NDCG@K scores of IFENet are closer to 1.0 compared to XGBoost scores for all $K$ values for all datasets. This indicates that the feature rankings produced by IFENet are more aligned with the ideal (or reference ranking) feature rankings compared to those produced by XGBoost.

IFENet enables a deeper analysis of how each feature affects the model's prediction, taking into account direct and indirect effects through interactions between features, while XGBoost offers a global view of feature importance that might overlook local interactions between features or feature contributions. Furthermore, the performance of IFENet in feature importance ranking is more stable for small and large datasets.

In addition, IFENET contains a built-in interpretability mechanism, which is inherently designed to provide explanations or insight into how it's making decisions for feature ranking using weights. Those weights are used during both the training and inference stages.

In addition, the attention scores assigned to each feature provide a direct measure of feature importance that is easily interpretable. This interpretability makes it easier to understand which features drive the model's predictions and why. Specifically, each feature has multiple importance scores that are calculated in each iteration. Each score expresses the feature importance of each iteration.

In summary, the result of the experiments demonstrates that IFENet more closely matches the reference ranking compared to the XGBoost ranking; IFENet obtains the highest NDCG scores, which means that IFENet produces a more focused and interpretable ranking compared to the XGBoost.

It is worth noting that while GradientSHAP provides a robust method for model-agnostic feature ranking, the reference ranking derived from it is inherently an interpretation relative to the underlying model and the SHAP methodology itself. Therefore, the ideal ranking in this context is understood as the ranking most closely aligned with the insights provided by GradientSHAP, rather than a universally objective or model-independent truth. This perspective adds important nuance to the evaluation of feature importance as used in some studies.

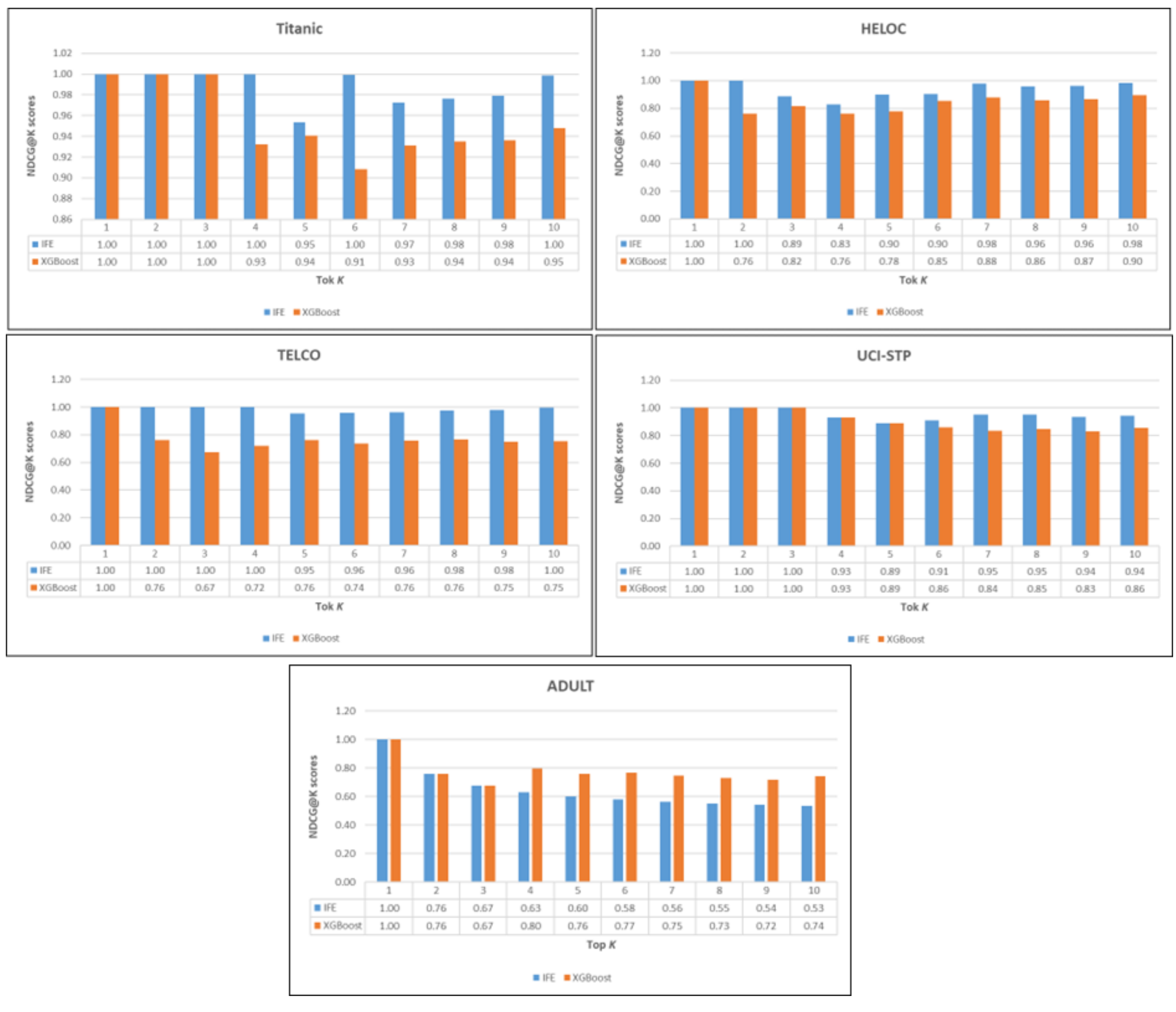

Figure 2. The results of Top-10 NDCG@K scores for IFENet and XGBoost with reference ranking according to GradientSHAP.

### 4.3.2 Baseline Model Comparison

This experiment aims to compare the classification performance of FNN with IFENet module with XGBoost, which is the baseline model for tabular data[37], [38]. The grid of hyperparameters is listed in Appendix A.1.

In experiments, IFENet outperforms XGBoost not only in feature importance classification but also in the classification task on most databases, albeit slightly (Table 3). On the Heloc, Titanic, and UCI-STP databases, IFENet achieves an improvement of 0.01 in F1-score compared with XGBoost. In terms of accuracy, IFENet outperforms XGBoost by 0.01 and 0.005 on the Heloc and Titanic datasets, respectively, while achieving comparable performance on the UCI-STP dataset.

On the TELCO database, IFENet outperforms XGBoost by 0.03 in F1-score, whereas XGBoost exceeds IFENet by 0.002 in accuracy. On the Adult dataset, XGBoost outperforms IFENet by 0.01 in F1-score and by 0.008 in accuracy.

Although the accuracy between XGBoost and IFENet is similar on the TELCO and UCI-STP datasets, IFENet achieves a higher F1-score. Furthermore, the proposed module improves the feature importance ranking, which can have a positive impact on the classification results.

Table 3. Comparison of classification performance between IFENet and XGBoost. Bold is the best result.

| | XGBoost | | IFENet | |
|---|---|---|---|---|
| Datasets | Accuracy | F1-score | Accuracy | F1-score |
| TELCO | **0.803** | 0.70 | 0.801 | **0.73** |
| HELOC | 0.738 | 0.74 | **0.749** | **0.75** |
| TITANIC | 0.795 | 0.77 | **0.800** | **0.78** |
| UCI-STP | **0.880** | 0.85 | **0.880** | **0.86** |
| ADULT | **0.864** | **0.81** | 0.856 | 0.80 |

### 4.3.3 Comparison with state-of-the-art models

In this experiment, the performance of IFENet model is compared with the state-of-the-art models in predicting tabular data. Four state-of-the-art models are considered: Neural Oblivious Decision Ensembles (NODE) [39], TabNet [40], Deep Abstract Networks (DANet) [41], and GANDALF[37].

NODE is a deep learning architecture consisting of differentiable oblivious decision trees (ODT) that are trained end-to-end by backpropagation and benefit from the power of multi-layer hierarchical representation learning.

TabNet is a novel deep learning architecture designed for tabular data, offering high performance and interpretability. It functions as an instance-based feature selection method, employing sequential attention to determine which features should be inferred at each decision stage.

DANet is a Deep Abstract Network for tabular data classification and regression, consisting of stacking such blocks of Abstract Layer (ABSTLAY). ABSTLAY block learns to explicitly group correlative input features and generate higher-level features for semantics abstraction.

Gandalf is a deep learning architecture that proposes a gating mechanism called the Gated Feature Learning Unit (GFLU) for feature selection. As mentioned in the related work section, these models show great performance in learning features in tabular data. Table 4 illustrates the results of the classification performance of the models.

It is worth noting that selecting tabular datasets for benchmarking is non-standard. Therefore, this evaluation aims to ensure that the performance of the proposed model is comparable, if not better than, the performance of other state-of-the-art models. As shown in Table 4, the performance of IFENet achieves superior performance in terms of accuracy and other evaluation metrics compared to the state-of-the-art models across all datasets except Titanic and Adult. For the Titanic dataset, the accuracy of IFENet is 0.005 lower than DANet, while its recall and F1-score measures are comparable to DANet.

Unlike the state-of-the-art models that require fine-tuning of numerous hyperparameters, the proposed module requires minimal tuning, which is $r$ value only.

As a summary, by iteratively excluding one feature at a time when measuring importance scores, the model eliminates that feature's influence and shifts focus to the remaining features. This process helps reveal hidden interactions and dependencies that might otherwise be obscured when all features are included. By observing how the model adapts to the absence of a particular feature, it can gain insights into how that feature interacts with other features and contributes to the overall predictive power of the model [42]. In addition, the proposed module allows for attributing a more precise attention score to the excluded feature in the next iteration, potentially revealing its true importance without interference from other features. Furthermore, the iterative process helps derive a comprehensive ranking of feature importance based on the observed impact on model performance[25], [42]. The feature ranking can be used to prioritize features for further analysis or to guide feature engineering efforts, ensuring that the most relevant features are utilized in subsequent modeling tasks.

Table 4. Comparison of classification performance between IFENet and the state-of-the-art models for all datasets. Bold is the best result.

| Model | TELCO | | | |
|---|---|---|---|---|
| | Accuracy | Precision | Recall | F1-score |
| NODE | 0.791 | 0.73 | 0.69 | 0.70 |
| TabNet | 0.785 | 0.73 | 0.65 | 0.67 |
| DANet | 0.791 | 0.73 | 0.70 | 0.71 |
| GANDALF | 0.795 | **0.74** | 0.70 | 0.72 |
| IFENet | **0.801** | **0.74** | **0.71** | **0.73** |
| | HELOC | | | |
| | Accuracy | Precision | Recall | F1-score |
| NODE | 0.735 | 0.74 | 0.74 | 0.73 |
| TabNet | 0.730 | 0.73 | 0.73 | 0.73 |
| DANet | 0.731 | 0.73 | 0.73 | 0.73 |
| GANDALF | 0.744 | **0.75** | **0.75** | 0.74 |
| IFENet | **0.749** | **0.75** | **0.75** | **0.75** |
| | TITANIC | | | |
| | Accuracy | Precision | Recall | F1-score |
| NODE | 0.769 | 0.79 | 0.73 | 0.74 |
| TabNet | 0.774 | 0.79 | 0.74 | 0.75 |
| DANet | **0.805** | 0.82 | **0.78** | **0.79** |
| GANDALF | 0.800 | **0.83** | 0.77 | 0.78 |
| IFENet | 0.800 | 0.80 | **0.78** | 0.78 |
| | UCI-STP | | | |
| | Accuracy | Precision | Recall | F1-score |
| NODE | 0.852 | 0.85 | 0.79 | 0.81 |
| TabNet | 0.861 | **0.89** | 0.79 | 0.82 |
| DANet | 0.828 | 0.80 | 0.79 | 0.80 |
| GANDALF | 0.861 | 0.86 | 0.81 | 0.83 |
| IFENet | **0.880** | 0.86 | **0.87** | **0.86** |
| | ADULT | | | |
| | Accuracy | Precision | Recall | F1-score |
| NODE | 0.852 | 0.81 | 0.78 | 0.80 |
| TabNet | **0.857** | **0.82** | **0.80** | **0.81** |
| DANet | 0.847 | 0.80 | 0.78 | 0.79 |
| GANDALF | 0.854 | 0.81 | **0.80** | 0.80 |
| IFENet | 0.856 | **0.82** | 0.79 | 0.80 |

#### 4.3.4 Efficiency Analysis

To assess the computational efficiency of the proposed module, we compare IFENet's training time, inference time, and the number of trainable parameters with other state-of-the-art models. For fairness, all models were run on the same hardware setup and with the same training and inference settings. The experiments are performed on the UCI-STD dataset with 64 batch size and 30 epochs of training. The parameter values for each model are chosen based on the best achieved accuracy, which are listed in Appendix II. The results of the experiments are reported in Table 5.

The training time includes the validation time, and the inference time excludes evaluation time. As shown in Table 5, our IFENet has the fewest number of parameters compared to the other models, and its inference time is comparable to the other models, although its training time is the longest. Although IFENet incurs a longer training time than some other models, this cost is justified given the improved interpretability and fine-grained feature selection it offers. Unlike conventional models that require post-hoc explanation tools (e.g., SHAP), IFENet learns feature importance inherently through its attention mechanism, providing built-in interpretability without additional explanation tools.

However, although IFENet has fewer hyperparameters, its iterative nature can be computationally expensive for large datasets. Specifically, the computational complexity of the IFE module is primarily driven by $d$ (number of features) forward passes through the attention unit. Each pass involves operations on the input data of size $n$ by $d$ where $n$ is the number of instances. The computational complexity can be approximated as $O(d.C_{attention})$, where $C_{attention}$ is the complexity of a single attention unit forward pass. This makes IFENet more computationally intensive than traditional attention (single-pass) mechanisms, especially when applied to datasets with a large number of features.

Table 5. The results of efficiency experiment.

| Model | Training Time (s) | Inference Time (ms/sample) | Parameters |
|---|---|---|---|
| NODE | 26.942 | 3.914 | 3.8M |
| TabNet | 29.911 | 0.584 | 70K |
| DANet | 228.233 | 2.809 | 2.8M |
| GANDALF | 34.092 | 0.617 | 2.9M |
| IFENet | 38.85 | 0.684 | 11.8K |

#### 4.3.5 Impact of amplification coefficient

The properties of the dataset influence the amplification coefficient, enabling the model to adjust the sensitivity of feature weight amplification. The classification results of the experiments are shown in Table 6 and Figure 3. The results demonstrate that the model's performance stabilizes when the value of $r$ exceeds 6.0. Additionally, the model's performance begins at a low level, increases until reaching a peak at a certain point, and then begins to decline. The peak performance varies across datasets but generally falls between one and six, as shown in Figure 3. The results indicate that the variation between values is not substantial. However, adjusting it provides a mechanism to control and refine classification accuracy according to the characteristics of each dataset.

Table 6. Impact of the amplification coefficient on classification accuracy.

| | Dataset | | | | |
|---|---|---|---|---|---|
| $r$ value | UCI-STP | HELOC | TELCO | TITANIC | ADULT |
| 1.0 | 0.708 | 0.744 | 0.798 | 0.774 | 0.858 |
| 2.0 | 0.847 | 0.742 | 0.798 | 0.785 | 0.858 |
| 3.0 | 0.876 | 0.743 | 0.798 | 0.8 | 0.856 |
| 4.0 | 0.871 | 0.741 | 0.797 | 0.79 | 0.855 |
| 5.0 | 0.861 | 0.742 | 0.796 | 0.8 | 0.854 |
| 6.0 | 0.861 | 0.743 | 0.796 | 0.79 | 0.854 |
| 7.0 | 0.861 | 0.749 | 0.795 | 0.79 | 0.854 |
| 8.0 | 0.861 | 0.741 | 0.798 | 0.79 | 0.857 |

#### 4.3.6 Impact of seed:

To verify the robustness of IFENet across different initialization seeds, experiments were repeated using three random seeds (1, 42, 3407[43]). Table 7 summarizes the corresponding results. The performance variations across seeds are minimal, indicating that IFENet produces stable and consistent outcomes.

Table 7. Impact of the seed on the IFENet performance in team of classification accuracy and F1-score.

| | Seed=1 | | Seed=42 | | Seed=3407 | |
|---|---|---|---|---|---|---|
| Dataset | Accuracy | F1-score | Accuracy | F1-score | Accuracy | F1-score |
| TELCO | 0.801 | 0.73 | 0.797 | 0.72 | 0.798 | 0.72 |
| HELOC | 0.749 | 0.75 | 0.748 | 0.75 | 0.743 | 0.74 |
| TITANIC | 0.800 | 0.78 | 0.779 | 0.76 | 0.800 | 0.79 |
| UCI-STP | 0.880 | 0.86 | 0.861 | 0.85 | 0.861 | 0.85 |
| ADULT | 0.856 | 0.80 | 0.858 | 0.81 | 0.858 | 0.80 |

Across different random seeds, the proposed model demonstrates consistent performance and stability. The accuracy variations are minimal, with differences not exceeding 0.004 across seeds for any dataset. Similarly, F1-scores remain nearly constant, varying by at most 0.02, confirming that the model's performance is robust to random initialization and not sensitive to stochastic factors.

#### 4.3.7 Ablation Study

In these experiments, the impact of the proposed ranking module on the simple classification model is evaluated. To do so, we compare the classification performance of the model with and without the proposed ranking module. The architecture of the simple predictive model is described and illustrated in Table 1. The hidden size layer is optimized as a hyperparameter, as shown in the Appendix.

Table 8 shows the classification results of both models for all datasets. We refer to FNN as IFENet without the ranking module (IFE module). As shown in Table 8, the proposed module improves the classification performance for all datasets. The proposed ranking module generates feature importance scores, applies weighting to the input features, and consequently improves classification performance. These results indicate that integrating the proposed ranking module makes the features more informative and leads to better classification outcomes.

Table 8. Impact of the proposed module on classification. Bold is the best result.

| | FNN | | IFENet | |
|---|---|---|---|---|
| Datasets | Accuracy | F1-score | Accuracy | F1-score |
| TELCO | 0.795 | 0.72 | **0.801** | **0.73** |
| HELOC | 0.743 | 0.74 | **0.749** | **0.75** |
| TITANIC | 0.785 | 0.77 | **0.800** | **0.78** |
| UCI-STP | 0.790 | 0.76 | **0.876** | **0.86** |
| ADULT | 0.847 | 0.79 | **0.856** | **0.80** |

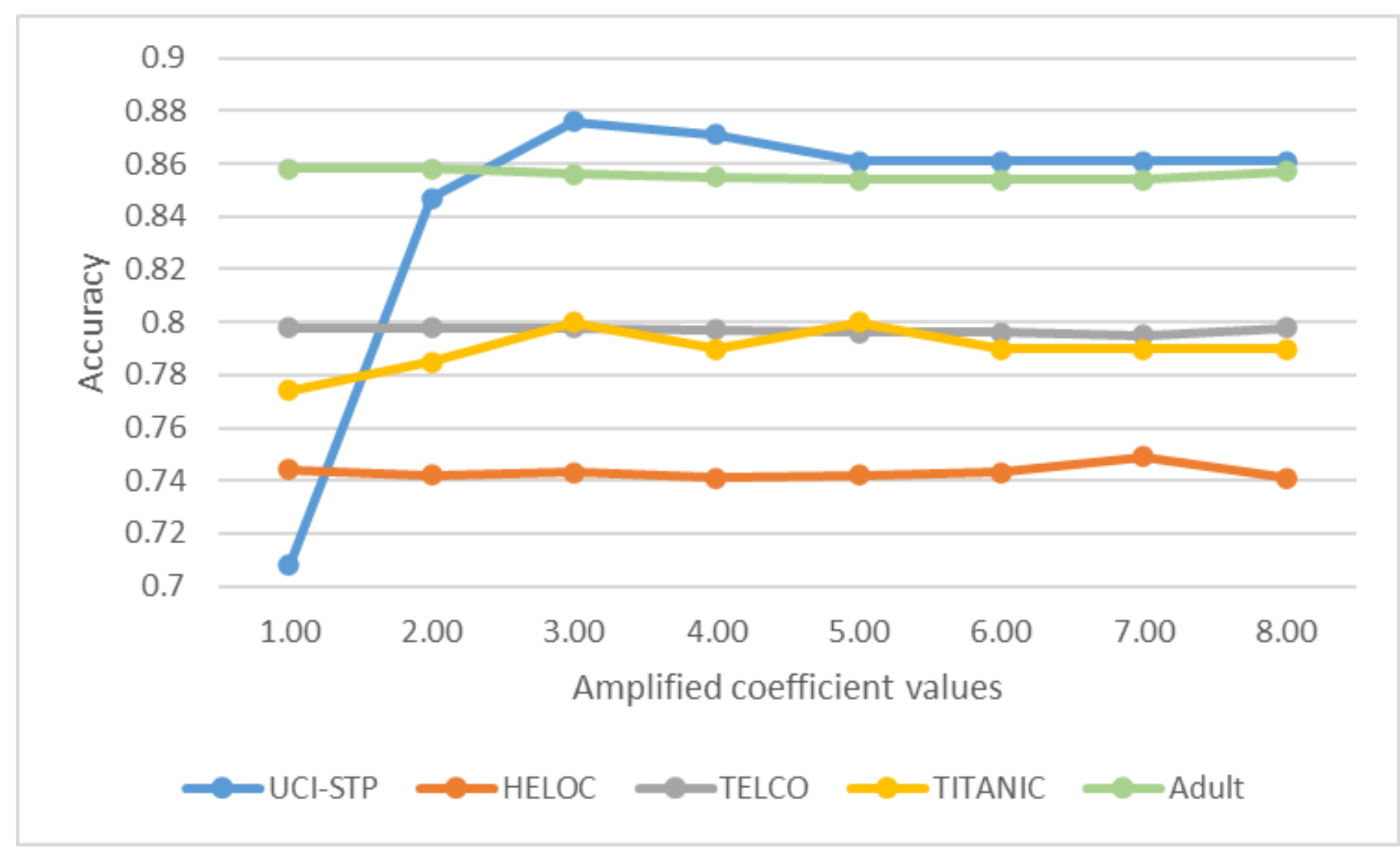

Figure 3. Impact of amplification coefficient for all datasets.

#### 4.3.8 Case Study: Prediction of Student Performance

We adopt the study conducted by Shaninah et al. [44] as a case study for predicting the performance of university students. The dataset is collected from undergraduate students at Zintan University in Libya through a structured survey distributed across various academic disciplines. The dataset has a total of 357 students with 56 features, including demographic, academic, and personality trait attributes. The class label consists of three categories based on the final grades: good, very good, and excellent. We refer to this dataset as Libyan University Student Performance (LUSP) dataset. Data preprocessing steps are applied to the dataset including cleaning and data encoding. In this experiment, we evaluate the proposed method in terms of classification and feature ranking performance. The classification accuracy and F1-score, and NDCG score are given in Table 9 and Figure 4 respectively. The results show that the proposed model achieved a robust and balanced performance both in the classification task and feature ranking.

Table 9. The performance of the proposed module on the LUSP dataset.

| Model | Accuracy | F1-score |
|---|---|---|
| IFENet | **0.952** | **0.95** |

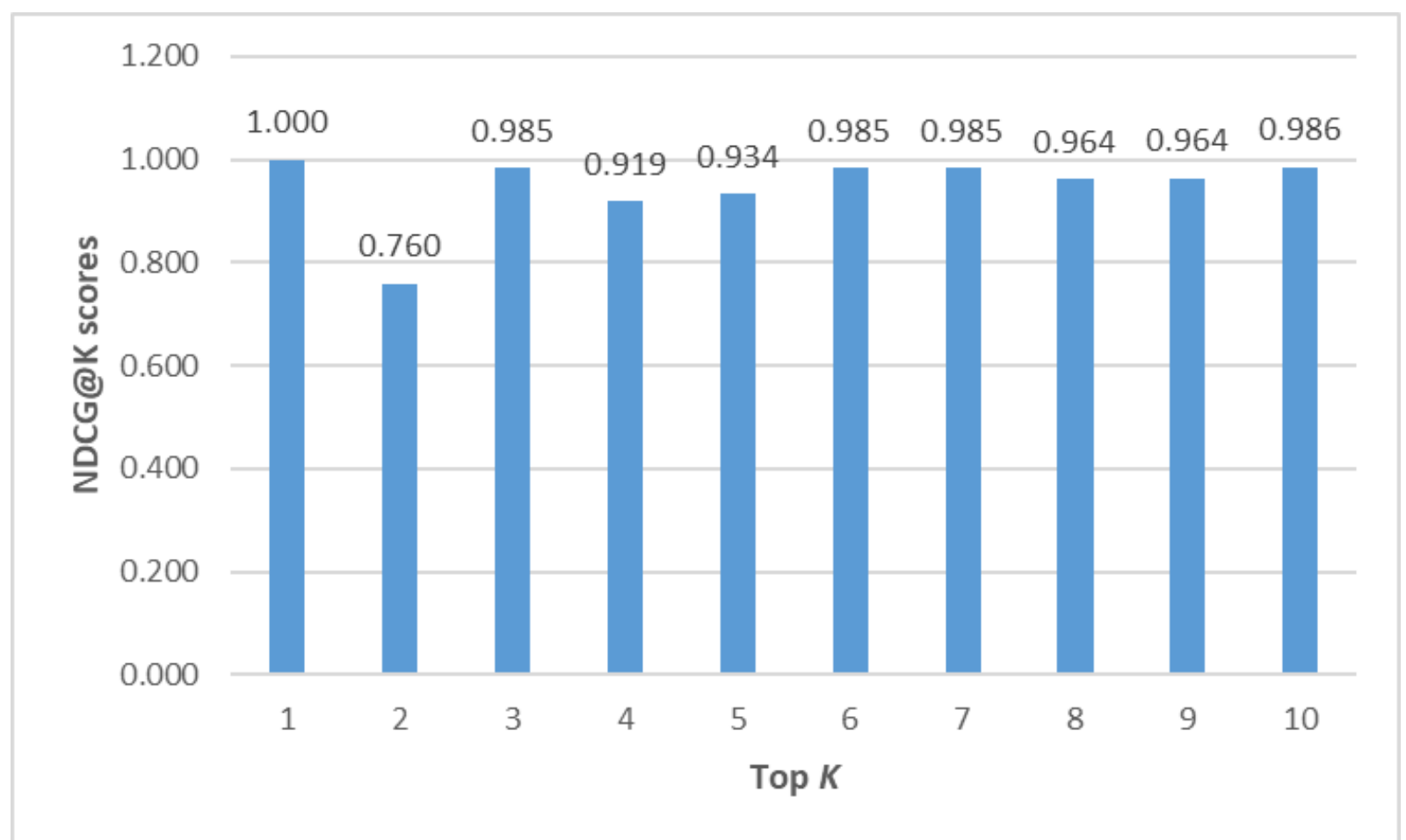

Figure 4. The results of Top-10 NDCG@K scores for IFENet with reference ranking according to GradientSHAP for LUSP dataset.

Table 10. Top 10 ranks of LUSP features by IFENet.

| Rank | Feature Number | Feature Type |
|---|---|---|
| 1 | 0 | Academic |
| 2 | 2 | Academic |
| 3 | 1 | Academic |
| 4 | 234 | Personality |
| 5 | 229 | Personality |
| 6 | 22 | Demographics |
| 7 | 174 | Personality |
| 8 | 189 | Personality |
| 9 | 23 | Demographics |
| 10 | 175 | Personality |

Table 10 lists the top 10 features ranked by the proposed method. As shown in the table, the academic features achieved the top ranking, followed by personality features, and then demographics features.

In summary, the robustness of the proposed model across multiple datasets is attributed to an iterative feature removal strategy, wherein the influence of each excluded feature on the residual feature set is systematically assessed.

Unlike traditional attention mechanisms, which typically assign attention weights to all features simultaneously, the IFENet iteratively assigns these weights by excluding a single feature in each iteration. This approach offers multiple advantages: it allows the model to adjust feature importance based on the current input context dynamically, mitigates bias from high-impact features, and reduces the risk associated with attention generalization limitations. In addition, it enables the model to reveal subtle relationships and interactions between features that might not be apparent when all features are considered at once. The proposed model repeatedly eliminates each feature from the input data in subsequent iterations. Whereas traditional attention mechanisms typically produce a single set of attention weights for all features, the proposed module aggregates the attention scores obtained from multiple iterations.

Here, it is worth noting that despite the proposed model having a superficial similarity to the RFE and sequential attention models, there are key differences between them. RFE is a wrapper-based method that iteratively removes features based on the performance of an external model, requiring retraining at each step. In contrast, IFENet integrates an iterative exclusion mechanism within the attention architecture, allowing the model to assess the relative contribution of each feature through differentiable attention recalibration, without retraining or permanently discarding inputs. Moreover, unlike sequential attention mechanisms, which distribute focus across layers or timesteps, IFENet explicitly leverages feature exclusion as a perturbation signal. By removing one feature per iteration, IFENet forces the model to redistribute its attention, thereby uncovering conditional dependencies and masked importance that are typically overlooked by single-pass or sequential attention mechanisms. This design enables IFENet to capture local and global feature relevance simultaneously, leading to more accurate and interpretable feature ranking.

While IFENet demonstrates strong performance across various tabular datasets, it is important to consider its potential limitations in certain scenarios beyond the cost that is already computed. For instance, in extremely high-dimensional datasets with hundreds or thousands of features, the iterative nature of the IFENet, even with its theoretical benefits, might introduce practical challenges related to memory footprint or convergence speed, particularly if feature dependencies become excessively complex or noisy. Similarly, for highly sparse tabular data, the repeated masking operations could potentially lead to unstable attention scores if the remaining features offer insufficient contextual information.

## 5. Conclusion

This study proposed a novel iterative feature exclusion module to enhance the understanding of feature importance in tabular data. Unlike traditional attention mechanisms, which typically assign attention weights to all features simultaneously, IFE iteratively assigns these weights by excluding one feature in each iteration. This approach enables the model to dynamically adjust feature importance based on the current input context, mitigating bias from high-impact features and reducing the risk associated with attention generalization limitations. For instance, in financial risk modeling, IFENet can reveal critical but conditional attributes influencing credit scoring or default prediction, thereby improving regulatory compliance and decision transparency. In medical diagnostics, the iterative exclusion process can uncover hidden interactions among clinical variables, enabling clinicians to understand how certain biomarkers gain importance only under specific conditions. Similarly, in industrial quality control and cybersecurity intrusion detection, IFENet can identify context factors contributing to anomalies, offering a foundation for interpretable AI systems in high-stakes environments. The proposed model's efficacy was evaluated on four public datasets. The results demonstrate that the module enhanced the features' importance ranking compared to XGBoost. In addition, the module outperforms state-of-the-art models for classification tasks.

The future work could explore adaptive iteration strategies or more robust attention mechanisms tailored for such challenging data characteristics to mitigate these potential issues, or investigate hybrid approaches that combine the IFE module with dimensionality reduction techniques for high-dimensional sparse data.

Furthermore, while IFENet is presented as a post-hoc interpretability module, its insights into feature importance present a clear path for future work aimed at model optimization and dimensionality reduction. We envision a potential extension where IFENet's analytical output is integrated directly into the training process, transforming it from a purely descriptive tool to a prescriptive one. One promising approach would be to incorporate a sparsity-inducing regularization term into the model's loss function. This term would be dynamically updated based on the feature importance scores derived from the IFENet attention mechanism.

## Acknowledgements

This work has been supported in part by the Ministry of Higher Education Malaysia for Fundamental Research Grant Scheme with Project Code: FRGS/1/2023/ICT02/USM/02/2.

## Statements and Declarations

- **Funding** This work was supported by the Ministry of Higher Education Malaysia (FRGS/1/2023/ICT02/USM/02/2).
- **Competing Interests** The authors declare that they have no known competing financial interests or personal relationships that could have appeared to influence the work reported in this paper.
- **Ethics approval** Not applicable.
- **Availability of Data and Materials** No data was generated for the research described in the article.

## Appendix I: Optimization of hyperparameters.

### IFENet

A list of IFENet hyperparameters along with their respective search spaces:

- Learning rate: Uniform choice {0.01, 0.001, 0.0001}
- Batch size: Uniform choice {32, 64, 128}
- Hidden size: Uniform distribution [16,128]
- r: Uniform distribution [1,5]

### XGBoost

A list of XGBoost hyperparameters along with their respective search spaces:

- Number of estimators: Uniform distribution [50, 300]
- Max depth: Discrete uniform distribution [3, 10]
- Min child weight: Uniform distribution [1,6]
- Subsample: Uniform distribution [0.5, 1]
- Colsample bytree: Uniform distribution [0.5, 1]
- Colsample bylevel: Uniform distribution [0.5, 1]
- Alpha: Log-Uniform distribution [0.5, 1]
- Gamma: Log-Uniform distribution [0.5, 1]

### NODE

A list of NODE hyperparameters along with their respective search spaces:

- Learning rate: Uniform choice {0.01, 0.001, 0.0001}
- Batch size: Uniform choice {32, 64}
- Number of layers: Uniform distribution [1,5]
- Number of Trees: Uniform choice {256,512}
- Additional Tree Dim: Uniform distribution [1,5]
- Depth: Uniform distribution [4,6]

### TabNet

A list of TabNet hyperparameters along with their respective search spaces:

- Learning rate: Uniform choice {0.01, 0.001, 0.0001}
- Batch size: Uniform choice {32, 64, 128}
- Prediction Dimension: Uniform distribution [4,60]
- Attention Dimension: Uniform distribution [4,60]
- N Step: Uniform distribution [3,10]

### DANet

A list of DANet hyperparameters along with their respective search spaces:

- Learning rate: Uniform choice {0.01, 0.001, 0.0001}
- Batch size: Uniform choice {32, 64, 128}
- N layers: Uniform choice {8,20,32}
- k: Uniform distribution [1,8]
- Abstlay dim 1: Uniform choice {16,32,64,128}
- Abstlay dim 2: Uniform choice {16,32,64,128}

### Gandalf

A list of Gandalf hyperparameters along with their respective search spaces:

- Learning rate: Uniform choice {0.01, 0.001, 0.0001}

- Batch size: Uniform choice {32, 64, 128}
- GFLU stages: Uniform distribution [3,20]
- GFLU dropout: Uniform distribution [0,0.9]

## Appendix II: The hyperparameter values used while calculating the minimum number of parameters:

**IFENet**

- Hidden size= 18
- r= 2.7182

**NODE:**

- Number of layers=3
- Number of Trees: Uniform choice=256
- Additional Tree Dim: Uniform distribution=1
- Depth=5
- Learning rate=0.01

**TabNet**

- Prediction Dimension=7
- Attention Dimension=38
- N Step=3
- Mask type='entmax'

**DANet**

- N layers=32
- K=5
- Abstlay dim 1=32
- Abstlay dim 2=32
- Learning rate=0.01

**Gandalf**

- GFLU stages=13
- GFLU_feature_init_sparsity=0.00549
- GFLU dropout: 0.449
- Learning rate=0.01